\documentclass[sigconf]{acmart} 
\AtBeginDocument{%
  }

\copyrightyear{2026}
\acmYear{2026}
\setcopyright{cc}
\setcctype{by}
\acmConference[MM '26]{Proceedings of the 34th ACM International Conference on Multimedia}{November 10--14, 2026}{Rio de Janeiro, Brazil}
\acmBooktitle{Proceedings of the 34th ACM International Conference on Multimedia (MM '26), November 10--14, 2026, Rio de Janeiro, Brazil}
\acmDOI{10.1145/3767308.3839017}
\acmISBN{979-8-4007-2213-4/2026/11}

\begin{document}

\title{CableDex: Cable Length Estimation on Industrial Reels Using a Handheld Device}

\author{Francisco Guillén}
\orcid{0009-0004-4710-3012}
\affiliation{%
  \institution{University of Beira Interior}
  \city{NOVA-LINCS UBI}
  \country{Portugal}}
\email{francisco.guillen@ubi.pt}

\author{Ricardo Almeida}
\orcid{0000-0000-0000-0000}
\affiliation{%
  \institution{COFICAB Portugal}
  \city{}
  \country{}}
\email{ricardo.almeida@coficab.com}

\author{Bruno M. C. Silva}
\orcid{0000-0002-5939-8370}
\affiliation{%
  \institution{University of Beira Interior}
  \city{IT - Instituto de Telecomunicações}
  \country{Portugal}}
\email{brunosilva@ubi.pt}

\author{João Neves}
\authornote{Corresponding author.}
\orcid{0000-0003-0139-2213}
\affiliation{%
  \institution{University of Beira Interior}
  \city{NOVA-LINCS UBI}
  \country{Portugal}}
\email{jcneves@ubi.pt}

\renewcommand{\shortauthors}{Guillén et al.}

\begin{abstract}
CableDex is a computer vision system that addresses the time-consuming and inaccurate manual measurement of cable length on industrial reels from a single photograph captured with a mobile phone. The system combines camera calibration, instance segmentation, pose estimation, and volumetric calculation to estimate the cable length across five different reel types and various cable sizes. This system is based on an instance segmentation model  trained on 1,000 manually annotated images, achieving 99.5\% mAP50 with an inference time of 5.66 ms per image. Evaluated on 75 reels across five reel types, the system achieves a MAPE of 4.90\%, within the 10\% error tolerance commonly accepted in industrial cable-reel measurement. The demonstration presents the end-to-end pipeline, from reel label scanning and image capture to segmentation and length estimation, through the mobile application. 

\end{abstract}



\maketitle
\section{Introduction}

The estimation of the length of a cable on industrial reels is a relevant problem in manufacturing environments. When cable is removed from a reel, operators cannot quickly determine the remaining length without specialized equipment or by completely unwinding it, making existing approaches either time-consuming or reliant on weight-based estimation. To address this problem, we introduce  CableDex, a computer vision system capable of estimating cable length from a single photograph acquired using hand-held devices. The first module delineates the reel and cable regions through an instance segmentation model. Then, a geometric information module estimates the intrinsic parameters of the camera by exploiting the known reel dimensions. Subsequently, the combination of these two modules allows recovering the pose of the reel in 3D space. Finally, the volumetric module estimates the volume occupied by the cable and the cable length is obtained by a volume/length ratio determined by the cable specifications.

\begin{figure}[b]
  \centering
  \includegraphics[width=1\linewidth]{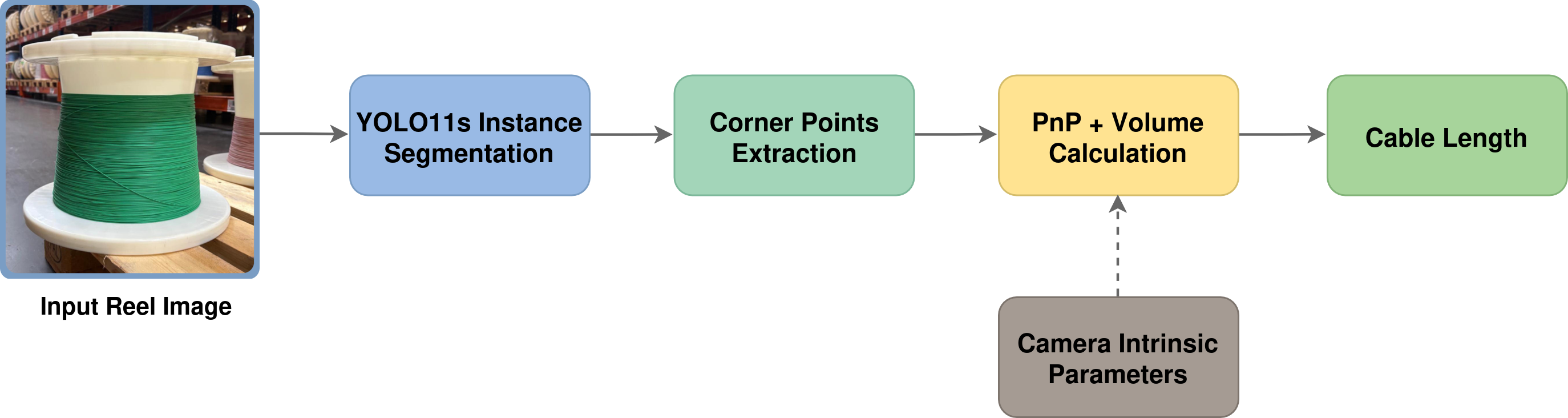}
  \caption{CableDex system pipeline: the input reel image is processed through instance segmentation and corner points extraction, combined with the camera intrinsic parameters for pose estimation and volume calculation, returning the estimated cable length.}
  \label{fig:pipeline}
\end{figure}

\section{System}

CableDex follows a client-server architecture consisting of three components: a Flutter~\cite{flutter} mobile application, a  Python/FastAPI~\cite{fastapi} backend server, and an external API for cable specifications and calibration data. The application handles image capture, reel label scanning, and result  visualization, while the computationally intensive processing runs on the backend. The backend receives the photograph along with the camera calibration parameters and reel identification data obtained  from the reel label scan, executes the computer vision pipeline, and returns the estimated cable length. The system pipeline is shown in Figure~\ref{fig:pipeline}.

\subsection{Camera Calibration}
Camera calibration estimates the intrinsic matrix $\mathbf{K}$ using Zhang's algorithm~\cite{zhang2000}, which can recover intrinsic parameters using only photographs of a flat pattern without specialized equipment. A centralized database stores calibration parameters organized by smartphone model, eliminating the need for repeated calibration. When a device is not registered, the user records a five-second video of a chessboard pattern (11$\times$7 internal corners), from which the backend extracts frames, detects corners with sub-pixel refinement, and computes the intrinsic matrix $\mathbf{K}$. The resulting parameters are stored both locally and in the database, making them available to other users of the same device model.

\subsection{Instance Segmentation}
A YOLO11s model~\cite{yolo11} segments the reel and cable regions, producing binary masks $B_{\text{reel}}$ and $B_{\text{cable}}$, trained on 1,000 manually annotated images covering five reel types in vertical and horizontal orientations, annotated in Roboflow~\cite{roboflow} using polygon segmentation, with data augmentation increasing the training set from 1,000 to 3,000 images. Only detections with confidence above 0.85 are accepted, with the reel and cable closest to the image center selected, achieving 99.5\% mAP50 and 99.48\% mAP50-95 with an inference time of 5.66~ms per image.

\subsection{Pose Estimation and Volume Calculation}

For the reel, the four extreme pixels of the segmentation mask are selected as the corner points. For the cable, each corner is determined as the closest pixel in $B_{\text{cable}}$ to the corresponding reel corner. The reel corners and their known 3D coordinates feed the Perspective-n-Point (PnP) algorithm, recovering the rotation matrix $\mathbf{R}$ and translation vector $\mathbf{t}$. The cable corners are then back-projected into 3D world coordinates, from which the upper radius $r_1$, lower radius $r_2$, and height $h$ of the cable region are determined. The cable volume is estimated using the truncated cone formula:

\begin{equation}
V_{\text{cable}} = \frac{\pi h}{3}(r_1^2 + r_1 r_2 + r_2^2)
\times (1 - \alpha)
\end{equation}
where $\alpha = 1 - \frac{\pi}{4} \approx 0.215$ accounts for the air fraction between circular cable turns. The cable length is then obtained by dividing $V_{\text{cable}}$ by the cable cross-section retrieved from the reel label.

\section{Demonstration}
The video starts with the mobile application loading screen, where the system automatically retrieves the camera calibration parameters for the device. The user then scans the reel label to obtain the reel details, which are used to retrieve the cable specifications from the product database. A photograph of the reel is captured and sent to the backend server, which segments the reel and cable regions, extracts the four corner points of each region (reel and cable), estimates the camera pose, and returns the estimated cable length. The result is displayed on the reel details screen, showing the reel description, identifier, estimated length, cross-section, and cable color. The user can then add the reel to the inventory, which stores the last 100 measured reels and can be exported as a CSV file.

\section{Experiments}
\subsection{Instance Segmentation}

Four YOLO11~\cite{yolo11} variants and Mask R-CNN~\cite{maskrcnn} were evaluated for instance segmentation (Table~\ref{tab:models}). YOLO11s was selected for its balance between accuracy and speed, achieving 99.5\% mAP50 and 99.48\% mAP50-95 with an inference time of 5.66~ms per image, six times faster than Mask R-CNN while offering comparable accuracy.

\vspace{-0.2cm}

\begin{table}[h]
\caption{Instance segmentation models comparison.}
\label{tab:models}
\small
\resizebox{0.48\textwidth}{!}{
\begin{tabular}{lrrrr}
\toprule
\textbf{Model} & \textbf{mAP50 (\%)} & \textbf{mAP50-95 (\%)} & \textbf{Train (min)} & \textbf{Infer (ms)} \\
\midrule
\textbf{YOLO11s} & \textbf{99.5} & \textbf{99.48} & \textbf{52.9} & \textbf{5.66} \\
YOLO11m & 99.5 & 99.50 & 104.3 & 11.01 \\
YOLO11l & 99.5 & 99.50 & 117.7 & 13.35 \\
YOLO11x & 99.5 & 99.50 & 208.6 & 24.88 \\
Mask R-CNN & 100.0 & 99.78 & 369.0 & 33.57 \\
\bottomrule
\end{tabular}
}
\end{table}
\vspace{-0.5cm}
\subsection{Cable Length Estimation}

The system was evaluated on 75 reels across five reel types, with 15 tests per type, achieving an overall Mean Absolute Percentage Error (MAPE) of 4.90\%, within the 10\% error tolerance commonly accepted in industrial cable-reel measurement.
All tests were conducted on-site at COFICAB Portugal's facilities under standard factory lighting conditions, with the complete process, from reel label scanning to the display of the estimated cable length.
Table~\ref{tab:length} reports the MAPE per reel type. The results show no correlation between reel size and estimation accuracy, suggesting that the main sources of error are not associated with reel dimensions but rather with image acquisition conditions, segmentation quality, or cable thickness.
\vspace{-0.2cm}
\begin{table}[h]
\caption{MAPE per reel type.}
\label{tab:length}
\begin{tabular}{lr}
\toprule
\textbf{Reel Type} & \textbf{MAPE (\%)} \\
\midrule
NPS150 & 3.92 \\
NPS250 & 6.03 \\
NPS400 & 4.52 \\
D600   & 4.22 \\
D800   & 5.83 \\
\midrule
\textbf{Average} & \textbf{4.90} \\
\bottomrule
\end{tabular}
\end{table}
\vspace{-0.5cm}

\section{Conclusion}
CableDex is a computer vision system for estimating the cable length on industrial reels from a single photograph using hand-held devices. The system combines camera calibration, instance segmentation, pose estimation, and volumetric calculation, achieving a MAPE of 4.90\% across five reel types, within the 10\% error tolerance commonly accepted in industrial cable-reel measurement. These results allow replacing manual and weight-based measurement approaches, representing a clear asset for manufacturing environments. Nevertheless, it should be noted that the truncated cone approximation assumes a regular winding pattern, and loosely or irregularly wound cable may deviate from this assumption, affecting estimation accuracy.

\section{Acknowledgments}
This work is funded by national funds through FCT – Fundação para a Ciência e a Tecnologia, I.P., and, when eligible, co-funded by EU funds under project/support UID/50008/2025 – Instituto de Telecomunicações, https://doi.org/10.54499/UID/50008/2025.
This work is also financed by the project WATERMARK\footnote{WATERMARK project (Watermark-Based Algorithms for Trustworthy Media Authentication and Robust Certification in Public Administration), Project No. 2024.07356.IACDC, supported by “RE-C05-i08.M04 – Support the launch of a program of R\&D projects aimed at the development and implementation of advanced systems in cybersecurity, artificial intelligence, and data science in public administration, as well as a scientific training program,” under the Recovery and Resilience Plan (PRR), as part of the funding agreement signed between the Recovery Portugal Task Force (EMRP) and the Foundation for Science and Technology (FCT).} and supported by UID/04516/NOVA Laboratory for Computer Science and Informatics (NOVA LINCS) with the financial support of FCT.IP.

\bibliographystyle{ACM-Reference-Format}
\bibliography{sample-base}

@String{Computing = "Computing" }

@String{Computer = "{IEEE} Computer" }

@ArtifactSoftware{R,
    title = {R: A Language and Environment for Statistical Computing},
    author = {{R Core Team}},
    organization = {R Foundation for Statistical Computing},
    address = {Vienna, Austria},
    year = {2019},
    url = {https://www.R-project.org/},
}

@misc{flutter,
  author = {Google},
  title = {Flutter - Build Apps for Any Screen},
  year = {2017},
  url = {https://flutter.dev}
}

@misc{fastapi,
  author = {Ram{\'i}rez, Sebasti{\'a}n},
  title = {FastAPI},
  year = {2018},
  url = {https://fastapi.tiangolo.com}
}

@article{zhang2000,
author = {Zhang, Zhengyou},
title = {A Flexible New Technique for Camera Calibration},
year = {2000},
publisher = {IEEE Computer Society},
volume = {22},
number = {11},
doi = {10.1109/34.888718},
journal = {IEEE Trans. Pattern Anal. Mach. Intell.},
month = nov,
pages = {1330--1334},
}

@software{yolo11,
  author = {Glenn Jocher and Jing Qiu},
  title = {Ultralytics YOLO11},
  version = {11.0.0},
  year = {2024},
  url = {https://github.com/ultralytics/ultralytics},
  orcid = {0000-0001-5950-6979, 0000-0003-3783-7069},
  license = {AGPL-3.0}
}

@misc{roboflow,
  author = {Dwyer, Brad and Nelson, Joseph},
  title = {Roboflow},
  year = {2020},
  url = {https://roboflow.com}
}

@misc{maskrcnn,
      title={Mask R-CNN}, 
      author={Kaiming He and Georgia Gkioxari and Piotr Dollár and Ross Girshick},
      year={2018},
      eprint={1703.06870},
      archivePrefix={arXiv},
      primaryClass={cs.CV},
      url={https://arxiv.org/abs/1703.06870}, 
}

\end{document}